\documentclass[conference]{IEEEtran}
\IEEEoverridecommandlockouts
\usepackage{cite}
\usepackage{amsmath,amssymb,amsfonts}
\usepackage{algorithm}
\usepackage{algorithmic}
\usepackage{graphicx}
\usepackage{textcomp}
\usepackage{xcolor}
\usepackage{subfigure}
\usepackage{epstopdf}
\usepackage{url}

\def\BibTeX{{\rm B\kern-.05em{\sc i\kern-.025em b}\kern-.08em
    T\kern-.1667em\lower.7ex\hbox{E}\kern-.125emX}}
\begin{document}

\title{Communication-Computation Efficient Device-Edge Co-Inference via AutoML
}

\author{\IEEEauthorblockN{Xinjie Zhang\IEEEauthorrefmark{1}, Jiawei Shao\IEEEauthorrefmark{1}, Yuyi Mao\IEEEauthorrefmark{2}, and Jun Zhang\IEEEauthorrefmark{1}}
\IEEEauthorblockA{
\IEEEauthorrefmark{1}Dept. of Electronic and Computer Engineering, The Hong Kong University of Science and Technology, Hong Kong \\
\IEEEauthorrefmark{2}Dept. of Electronic and Information Engineering, The Hong Kong Polytechnic University, Hong Kong \\
Email: \{xinjie.zhang, jiawei.shao\}@connect.ust.hk, yuyi-eie.mao@polyu.edu.hk, eejzhang@ust.hk}
}

\maketitle
\begin{abstract}
Device-edge co-inference, which partitions a deep neural network between a resource-constrained mobile device and an edge server, 
recently emerges as a promising paradigm to support intelligent mobile applications. To accelerate the inference process, on-device 
model sparsification and intermediate feature compression are regarded as two prominent techniques. However, as the on-device 
model sparsity level and intermediate feature compression ratio have direct impacts on computation workload and communication overhead
respectively, and both of them affect the inference accuracy, finding the optimal values of these hyper-parameters brings 
a major challenge due to the large search space. In this paper, we endeavor to develop an efficient algorithm to determine these hyper-parameters. 
By selecting a suitable model split point and a pair of encoder/decoder for the intermediate feature vector, this problem is casted as 
a sequential decision problem, for which, a novel automated machine learning (AutoML) framework is proposed based on 
deep reinforcement learning (DRL). Experiment results on an image classification task demonstrate the effectiveness of the proposed 
framework in achieving a better communication-computation trade-off and significant inference speedup against various baseline schemes.
\end{abstract}

\begin{IEEEkeywords}
Device-edge co-inference, deep neural network (DNN), automated machine learning (AutoML), deep reinforcement learning (DRL), 
communication-computation trade-off.
\end{IEEEkeywords}

\section{Introduction}
The past decade has witnessed the remarkable success of deep neural networks (DNNs) in a large variety of applications. Unfortunately, DNN-based 
applications are generally computation-intensive, which makes mobile devices with limited computational resources incapable of providing  
timely and reliable inference services. Mobile edge computing (MEC), which injects Cloud Computing capabilities into the wireless network edge,
brings new possibilities of achieving low latency mobile intelligence \cite{b1}. With the aid of MEC, DNN models can be 
deployed at an edge server with relatively abundant computational resources, and thus mobile devices can offload their raw data for 
server-based inference \cite{b2}.

Nevertheless, server-based inference might incur significant communication overhead especially for applications with large data dimension, 
e.g., 3D point cloud classification. Fortunately, the fast-evolving chip
technologies give birth to advanced mobile processors, empowering mobile devices to handle lightweight DNN processing. As a result, 
device-edge co-inference, where a DNN is partitioned between a mobile device and an edge server, emerges as a promising 
solution to avoid offloading raw data from mobile devices \cite{c1}. In particular, for each inference request, a mobile device first 
processes the on-device DNN partition, and transmits an intermediate feature vector to the edge server. The edge server then uses the received 
intermediate feature vector as input of the server DNN partition for further processing, and feeds back the inference result. 
As the server DNN partition usually demands higher computations compared to the on-device counterpart, device-edge co-inference is 
effective in balancing the on-device computation workload and communication overhead. 

While device-edge co-inference was proposed for reducing the communication overhead, it is inadequate for achieving low-latency inference 
in practice due to the in-layer data amplification phenomenon in many popular DNN models \cite{b3}. Specifically, dimensions of the 
intermediate feature vectors of early neural network layers even exceed that of the raw data, so the network can be split only at later 
layers to avoid a too high communication overhead, which shall increase the on-device computation workload, and deplete the merits of 
device-edge co-inference. To resolve this dilemma, preliminary attempts 
introduced model sparsification and feature compression techniques for edge inference \cite{b4,b5,d1,b6,b7}. In \cite{b4}, a two-step 
pruning framework that integrates model splitting with convolutional filter pruning was proposed in order to reduce both the communication 
and computation workload. To relieve the adverse impacts of data amplification, a learning-based end-to-end architecture 
was developed for efficient feature compression and transmission for image classification \cite{b5} and point cloud processing \cite{d1}. 
Besides, model pruning and feature encoding techniques were jointly utilized for collaborative inference over noisy wireless channels in \cite{b6}. 
In addition, a three-step framework based on model splitting, communication-aware model compression, and task-oriented feature encoding, were 
proposed in \cite{b7}, and the critical communication-computation trade-off in device-edge co-inference systems was investigated.

However, on one hand, the hyper-parameters, including the model sparsity level and intermediate feature compression ratio, were obtained by 
manual adjustment for different model split points in prior studies, which is laborious and time-consuming due to the large search space. On the other hand,
existing schemes perform on-device model sparsification and intermediate feature compression independently, which neglect
their tight couplings on communication overhead, on-device computation workload, and inference accuracy, and thus may result in low-quality
solutions. These necessitate an efficient algorithm to jointly optimize the on-device model sparsity level and intermediate feature 
compression ratio, meanwhile, taking the potential inference accuracy degradation into considerations. 

In this paper, we propose an automated machine learning (AutoML) framework to achieve communication-computation efficient device-edge co-inference 
based on deep reinforcement learning (DRL), which determines the sparsity level for each on-device DNN layer and the compression ratio for 
the intermediate feature vector. By selecting a suitable model split point for a backbone DNN model and inserting a pair of intermediate 
feature encoder/decoder, we develop a deep deterministic policy gradient (DDPG) algorithm to train an agent that automatically prunes 
unimportant filters to an optimized sparsity level for each on-device network layer using the one-shot filter pruning method \cite{b8}, 
and simultaneously devise a lightweight autoencoder for feature compression. We compare the proposed AutoML framework against various 
existing edge inference schemes via numerical experiments. Our results show that the proposed framework reduces up to 87.5\% of the
communication overhead and saves 70.4\% of the on-device computations with less than 1\% accuracy loss compared with server-based inference 
and simple model partition, respectively. In addition, it achieves a better communication-computation trade-off 
and enjoys significant end-to-end inference speedup than existing schemes.

\begin{figure*}[htbp]
    \centering
    \includegraphics[height=7.9cm,width=14.91cm]{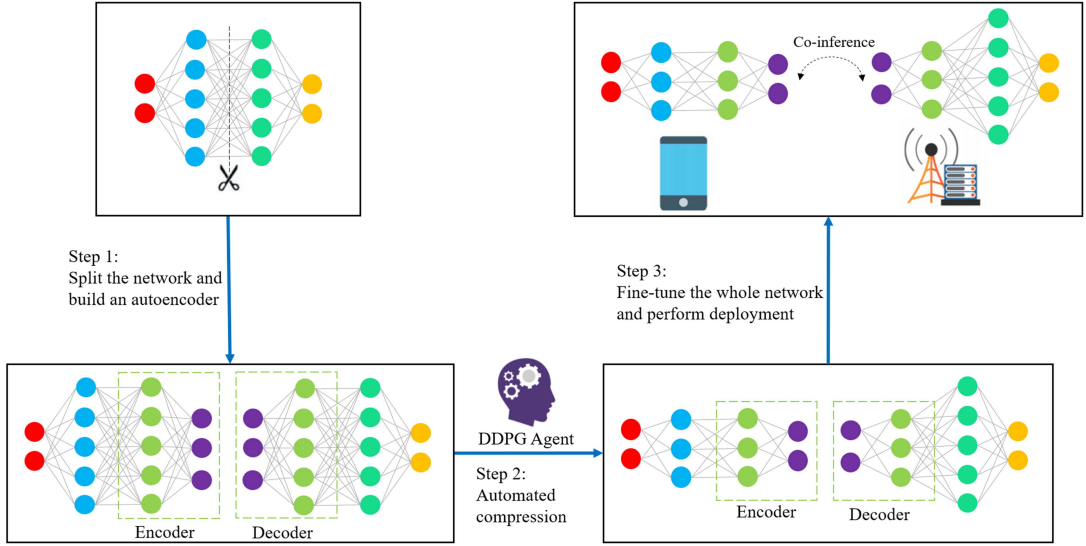}
    \caption{The proposed AutoML framework for device-edge co-inference. \textbf{Step 1:} We select a split point to partition 
    a pretrained DNN and insert an autoencoder for rudimentary compression of the intermediate feature vector. \textbf{Step 2:} Both the on-device 
    model and the autoencoder are compressed by heterogeneous filter pruning. We apply DRL to automatically search the optimal model sparsity 
    level for each on-device layer and a compressed intermediate feature autoencoder. \textbf{Step 3:} We fine-tune 
    the entire network to improve the inference accuracy and deploy the two partitions at a mobile device and an edge server, respectively.}
    \label{fig:AutoML framework}
\end{figure*}

\section{Preliminaries} \label{sec:preliminary}
In this section, we first introduce the background on model sparsification and feature compression, which are enabling techniques for
communication- and computation-efficient device-edge co-inference. Recent advancements on AutoML are then briefly reviewed.

\subsection{Model Sparsification and Feature Compression}
Model sparsification, which prunes unimportant network parameters to reduce computation and memory requirements, is one of the most effective 
techniques to accelerate DNN processing \cite{c3}. There are two types of model pruning techniques, namely, unstructured 
pruning \cite{c4} and structured pruning \cite{c5}. For unstructured pruning, each redundant parameter is pruned independently, which admits 
a high compression ratio. However, inference acceleration is difficult to achieve without specialized hardware due to the resulting irregular 
sparsity patterns. In contrast, structured pruning, which induces regular sparsity patterns by pruning the entire weight tensors, 
is able to boost DNN processing with off-the-shelf hardware, and thus more preferable for device-edge co-inference.

In parallel, feature compression reduces the amount of data that needs to
be transmitted from mobile devices to the edge server for collaborative inference. While traditional hand-crafted data compression algorithms 
were utilized for feature compression \cite{b3}, they are primarily designed for data recovery and fail to exploit
characteristics of the inference tasks. Hence, learning-based feature compression (a.k.a. feature encoding) algorithms have recently received 
significant interests \cite{b5,d1,b6,b7,b10}, which facilitates automatic discovery of task-irreverent information so that the 
communication overhead can be reduced more effectively. 

However, existing feature compression algorithms ignore the complex interplay with model sparsification, which motivates a joint 
consideration as will be pursued in this work. 

\subsection{AutoML}
AutoML, which is often referred to the process of automating machine learning model developements, has been applied to many important 
problems in deep learning \cite{b11}. Most studies on AutoML for DNN inference acceleration mainly focused on on-device inference 
and server-based inference \cite{b12,b13}. 
For instance, a network to network compression algorithm was developed based on policy gradient reinforcement learning in \cite{b12}. 
Considering the dependence among different DNN layers, the optimal model compression rate for each layer was obtained based on DRL 
in \cite{b13}. This study was extended for device-edge co-inference in \cite{b14}, which leverages a DRL algorithm to determine the sparsity level for each DNN layer 
based on the feedback of the hardware accelerator and system status. However, simply pruning the last on-device network layer 
cannot completely eliminate the negatives of data amplification. As a result, it calls for a novel AutoML framework combining model sparsification with 
feature compression in order to achieve communication-computation efficient device-edge co-inference.

\section{A DRL-based AutoML Framework for Efficient Edge Inference} \label{sec:proposed framework}
This section first gives an overview of the proposed AutoML framework. The task of determining the optimal 
model hyper-parameters is formulated as a sequential decision problem, for which, a DDPG algorithm is developed.

\subsection{Overview} 
As shown in Fig. \ref{fig:AutoML framework}, the proposed AutoML framework consists of three main steps. First, we select a split point 
at a pretrained DNN model, insert and train a feature autoencoder composed of a pair of complementary encoder and decoder for rudimentary 
intermediate feature compression. In particular, the encoder is made up of a convolutional layer, which shrinks the feature map by half 
in width and height, and removes 7/8 of the channels, as well as a fully-connected layer that reduces 3/4 of the intermediate feature dimension. 

Next, we utilize the heterogeneous filter pruning method developed in \cite{b8} for model sparsification and feature compression, which 
is a magnitude-based one-shot pruning approach that determines the importance of the filters according to their $l_1$-norms. 
It prunes filters in convolutional layers to reduce the amount of on-device computations, meanwhile, removes neurons in the 
fully-connected layer of the intermediate feature encoder for further communication overhead reduction. Pruning of the 
intermediate feature decoder uses the same compression ratio as that of the encoder due to their complementary symmetry. 
As will be detailed in the next subsection, in order to obtain the optimal sparsified on-device model and a lightweight autoencoder, 
DRL is applied to search for the model sparsity level for each on-device network layer and the compression ratio for the intermediate 
feature vector. Note that in contrast to \cite{b14}, the server partition is not sparsified since it severely degrades the inference accuracy with marginal 
additional latency reduction given abundant computational resources at the edge server.
 
In the last step, we fine-tune the entire network to further improve the accuracy before deployment. 

\subsection{Problem Formulation and Its Key Elements} \label{subsec:problem formulation}
We formulate the task of determining the on-device model sparsity level and intermediate feature compression ratio as a sequential decision 
problem with its key elements defined as follows.

\subsubsection{State Space}
The system state consists of 12 components for each layer, which distinguishes different DNN layers.
It can be written for layer $L_i$ as follows:
\begin{equation}
    \begin{aligned}
        s_i = & (i, type_i, k_i, stride_i, c^{in}_i, c^{out}_i, f^{in}_i, \\
             & FLOPs_i, reduced_i, rest_i, d_i, a_{i-1}),  
    \end{aligned}    
\end{equation}
where $i$ is the layer index, $type_i$ denotes the type of layer $L_i$ (i.e., a convolutional layer or fully-connected layer), and $k_i$ 
represents the kernel size. $c^{in}_i$ and $c^{out}_i$ are the numbers of input and output channels, respectively. For simplicity, we assume 
that the height and width of the feature map are identical, denoted as $f^{in}_i$. Besides, $FLOPs_i$ denotes the amount of 
floating point operations (FLOPs) required to process layer $L_i$, $reduced_i$ denotes the amount of reduced FLOPs in previous layers, 
and $rest_i$ represents the total amount of FLOPs in the remaining layers. In addition, $d_i$ is the transmitted data size, which is 
positive for the last layer of the feature encoder and zero for other layers, and $a_{i-1}$ denotes the action for layer $L_{i-1}$. 
We normalize each element in the state tuples to $[0, 1]$ for ease of decision making.

\subsubsection{Action Space}
We define the preserved ratio $a_i$ as the action for layer $L_i$, which equals 1 minus the prune rate and determines the sparsity level of 
each on-device layer (compression ratio of intermediate feature vector). Similar to \cite{b13}, we let $a_i \in (0, 1]$ to
achieve fine-grained compression. 

\subsubsection{Reward Function}
To evaluate the resulting pruned on-device model and compressed intermediate feature vector, we slightly modify the macro $F_1$-score 
formula \cite{b15} to define an innovative reward function as follows:
\begin{equation}
    \begin{aligned}
        R \triangleq \frac{R_1+R_2+\beta R_3}{3}, \label{eq:reward function}
    \end{aligned}
\end{equation}
where $R_1 \triangleq \frac{2\kappa\nu}{\kappa+\nu}$, $R_2 \triangleq \frac{2\kappa\rho}{\kappa+\rho}$, and 
$R_3 \triangleq \frac{2\nu\rho}{\nu+\rho}$. We denote the inference accuracy of the pruned model as $\kappa$, and let 
$\nu \triangleq 1-\frac{\lambda}{\Lambda}$ and $\rho \triangleq 1-\frac{\omega}{\Omega}$ be the sparsity level of the entire on-device model and
the intermediate feature compression ratio, respectively. In these expressions, $\lambda$ ($\omega$) is the amount of FLOPs required to 
process the on-device model partition (size of the intermediate feature vector) after model pruning (intermediate feature compression) and 
$\Lambda$ ($\Omega$) is the value corresponds to the original model. The term $R_1$ balances the inference accuracy and model sparsity, 
and $R_2$ and $R_3$ are similarly defined. Since the inference accuracy of the pruned network is sensitive to the sparsity level of the on-device model 
as well as the feature compression ratio \cite{b13}, while $R_3$ is independent of the inference accuracy, we introduce a weighting factor $\beta \in [0,1]$ to 
avoid having a highly-compressed model with low inference accuracy. Such a reward function also generalizes the one adopted in \cite{b14} that only considers the latency 
reduction achieved by model pruning.

\begin{algorithm}[tbp]
    \caption{DDPG for Device-edge Co-inference}
    \label{alg:DDPG}
    \begin{algorithmic}[1]
    \REQUIRE ~~ \\ 
    Randomly initialize an online actor network $\mu$ and online critic network $Q$ parameterized by $\theta^\mu$ and $\theta^Q$, respectively;
    Initialize a target actor network $\mu'$ and target critic network $Q'$ parameterized by $\theta^{\mu'} \leftarrow \theta^\mu$ and $\theta^{Q'} \leftarrow \theta^Q$, respectively;
    Set $\mathcal{B} \leftarrow \emptyset$ and $R_{opt} \leftarrow 0$.
    \ENSURE ~~\\ 
    The optimized preserved ratios $\{a_i^{opt}\}$, $i$ = 1, ..., MaxLayer.
  
    \FOR {$episode$ = 1:MaxEpisode}
        \FOR{$t$ = 1:MaxLayer}
            \STATE{Observe the system state $s_t$ and select the preserved ratio according to $a_t=clip(\mu^n(s_t))$.}
            \STATE{Compress layer $L_t$ using the one-shot filter pruning method based on $a_t$.}
        \ENDFOR
        \STATE{Evaluate the inference accuracy of the compressed model with slight fine-tuning on a validation dataset
        and obtain the episode reward $R_{episode}$ as defined in (\ref{eq:reward function}).}
        \FOR{$t$ = 1:MaxLayer}
            \STATE{$r_t \leftarrow R_{episode}$.}
            \STATE{Store $(s_t,a_t,r_t,s_{t+1})$ in the replay buffer $\mathcal{B}$.}
            \IF{the warm-up phase is completed}
                \STATE{Randomly select $N$ samples from the buffer $\mathcal{B}$.}
                \STATE{Update the online critic and actor networks by optimizing (\ref{eq:loss function}) using the Adam algorithm.}
                \STATE{Update the target actor and critic networks according to: \\
                    \qquad\qquad $\theta^{\mu'}\leftarrow \tau\theta^\mu+(1-\tau)\theta^{\mu'}$ \\
                    \qquad\qquad $\theta^{Q'}\leftarrow \tau\theta^Q+(1-\tau)\theta^{Q'}$.}
            \ENDIF
        \ENDFOR
        \IF{$R_{episode} \geq R_{opt}$ }
        \STATE{Set $R_{opt}\leftarrow R_{episode}$ and $\{a_i^{opt}\}\leftarrow \{a_i\}$.}
    \ENDIF
    \ENDFOR
    \end{algorithmic}
    \end{algorithm}

\addtolength{\topmargin}{0.1in}
\subsection{The DDPG Algorithm}
The DDPG algorithm is adopted to choose actions from a continuous space due to its less reliance on a large
number of training samples and good generalizability to large state space \cite{b16}.
In particular, we train a DDPG agent based on an actor-critic architecture, where an actor network and a critic network are utilized to 
approximate the policy and value functions, respectively \cite{b17}. The input and output of the actor network are the state
and action, respectively, while the critic network determines a value for each state-action pair. The training process of the DDPG agent 
consists of a warm-up phase and an update phase. In the warm-up phase, we employ a replay buffer $\mathcal{B}$ to store sufficient state 
transitions as training samples without updating the agent. When the number of the training samples reaches a certain thresold (which is 
2/3 of the replay buffer size in our experiments), the training process enters the update phase, in which, training samples are randomly 
drawn from the replay buffer to optimize the agent. 

As shown in Algorithm \ref{alg:DDPG}, we train the DDPG agent in an episodic style, where each episode solves the sequential decision problem 
defined in Section~\ref{subsec:problem formulation} once. We denote MaxEpisode and MaxLayer as the maximum number 
of episodes and the number of network layers needed to be pruned/compressed, respectively. For each layer, we take state $s_t$ as input of 
the online actor network, which outputs a preserved ratio $a_t$. To avoid the state space being trapped in a local minima of the reward 
function, we construct an exploration policy as $\mu^n(s_t)\thicksim TN(\mu(s_t|\theta_t^{\mu}), \sigma^2, 0, 1)$, where 
$TN(\mu, \sigma^2, \alpha, \gamma)$ denotes a normal distribution with mean $\mu$ and variance $\sigma^2$ truncated to the range of 
$[\frac{\alpha-\mu}{\sigma}, \frac{\gamma-\mu}{\sigma}]$. A clipping function $clip(\cdot)$ is used to restrict $a_t$ within $\left(0,1\right]$. 
Based on the value of the preserved ratio $a_t$, we execute the one-shot filter pruning method for layer $L_t$ and the system then transits 
to the next state $s_{t+1}$ corresponding to layer $L_{t+1}$. After compressing the fully-connected layer of the intermediate feature encoder, 
we evaluate the inference accuracy of the compressed model with slight fine-tuning on a validation dataset, and obtain the episode 
reward $R_{episode}$ as defined in (\ref{eq:reward function}). The value of $R_{episode}$ is used as the reward $r_t$ for each action in the 
current episode, and the tuple $(s_t,a_t,r_t,s_{t+1})$ is stored as a training sample in the replay buffer $\mathcal{B}$.

After completing the warm-up phase, we randomly draw $N$ samples from the replay buffer to train the online critic 
and actor networks by minimizing their respective loss functions $J(\theta^Q)$ and $J(\theta^{\mu})$ \cite{b16} using 
Adam optimizer \cite{c7} as follows:  
\begin{equation}
    \begin{aligned}
        J(\theta^Q) = \frac{1}{N}\sum_{j}(y_j-Q(s_j, a_j|\theta^Q))^2, \\
        J(\theta^{\mu}) = -\frac{1}{N}\sum_{j}Q(s_j,\mu(s_j|\theta^\mu)|\theta^Q), \label{eq:loss function}
    \end{aligned}
\end{equation}
where $y_j \triangleq r_j - b + Q'(s_{j+1}, \mu'(s_{j+1}|\theta^{\mu'})|\theta^{Q'})$ and $b$ is an exponential 
moving average of the mean of the previous batch rewards. The target networks are then optimized by soft updates with a small updating rate 
$\tau$ to ensure stable training. The series of actions $\{a_i^{opt}$\} with the maximal sepisode reward $R_{opt}$ is output as the final solution 
of the sequential decision problem for model sparsification and feature compression.

\begin{figure}[t]
    \centering
    \includegraphics[height=6.04cm,width=8.18cm]{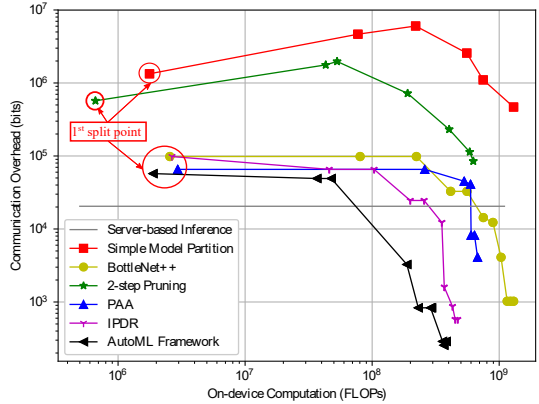} 
    \caption{The communication-computation trade-off curves for different edge inference schemes.}
    \label{fig:communication-computation trade-off}
\end{figure}

\section{Experiment Results} \label{sec:experiment}
\subsection{Experiment Setup}
In order to evaluate the performance of the proposed AutoML framework, we consider an image classification task on the CIFAR-10 dataset \cite{c8}, 
which is composed of 60,000 color images of 32$\times$32 pixels in 10 classes. These images are divided into a training set, a validation set, 
and a test set with 45,000, 5,000, and 10,000 images, respectively. We use ResNet-50 \cite{b18} for image classification. Specifically, for device-edge 
co-inference, we split the ResNet-50 and apply the proposed AutoML framework for different model split points. Note that not all 
layers in ResNet-50 are suitable split points because of the shortcut connection structure. Thus, we regard each residual block in 
ResNet-50 as a candidate split point.  

In our experiments, the original ResNet-50 attains 92.99\% classification accuracy. We use FLOPs to approximate the amount of 
on-device computations and adopt the transmitted data size as a measurement of the communication overhead. Both the actor and critic networks 
in Algorithm \ref{alg:DDPG} consists of two hidden layers with 300 neuros \cite{b13}. We set MaxEpisode and $\tau$ as 1100 and 0.01, respectively. 
The DDPG agent is trained with 64 as the batch size. The learning rates for the actor and 
critic network are 0.001 and 0.0001, respectively. 

\subsection{Baseline Schemes}
We adopt the following baseline edge inference schemes for comparsion:
\begin{figure*}[t]
    \centering
    \subfigure[Raspberry Pi 3B+.] 
    {\label{subfig:Raspberry pi}\includegraphics[height=6.04cm,width=8.18cm]{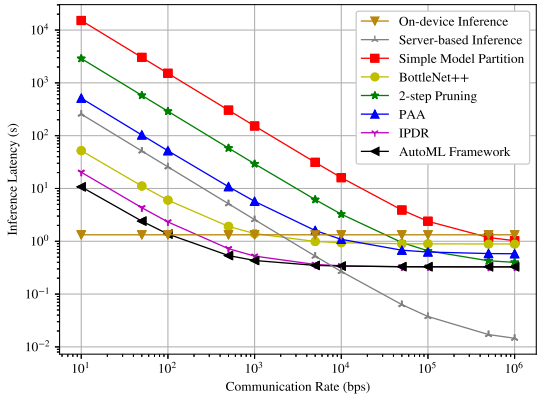}}\hspace{7mm} 
    \subfigure[Honor 8 Lite smartphone.] 
    {\label{subfig:Honor lite} \includegraphics[height=6.04cm,width=8.18cm]{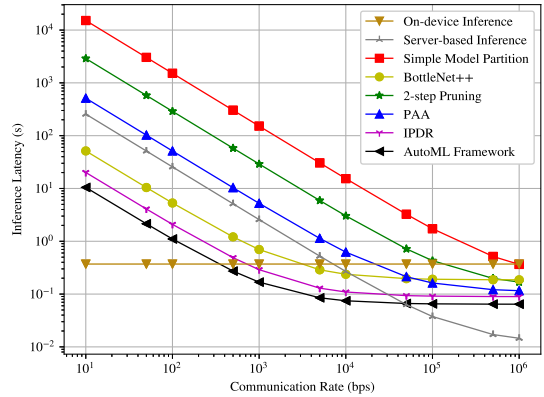}}\hspace{7mm} 
    \caption{End-to-end inference latency vs. communication rate.}
    \label{fig:latency analysis}
    \end{figure*}
\begin{enumerate}
    \item \textbf{Server-based Inference}: A pretrained ResNet-50 is deployed at the edge server and the mobile device transmits
    the PNG-compressed images to the edge server for inference. 
    \item \textbf{Simple Model Partition}: A pretrained ResNet-50 is partitioned between a mobile device and an edge server, and the 
    intermediate feature vector is compressed with Huffman coding. 
    \item \textbf{BottleNet++} \cite{b5}: This scheme is similar to simple model partition, except that a trainable autoencoder 
    replaces Huffman coding for more efficient data compression. 
    \item \textbf{2-step Pruning} \cite{b4}: This is a device-edge co-inference scheme that prunes the on-device model partition via two steps: 
    The first step prunes the entire pretrained model whereas the second step only prunes the layer right before the split 
    point. Huffman coding is used to encode the intermediate feature vector.
    \item \textbf{Pruning + Asymmetrical Autoencoder (PAA)} \cite{b6}: PAA splits the ResNet-50 into two partitions for device-edge co-inference, 
    where the on-device partition is pruned and the intermediate feature vector is compressed by an asymmetrical autoencoder.
    \item \textbf{Incremental Pruning + Dimension Reduction (IPDR)}: Different from PAA, IPDR uses incremental pruning to compress the
    on-device model partition together with the first step of task-oriented encoding \cite{b7} to reduce the dimensions 
    of the intermediate feature vector. 
\end{enumerate}

\subsection{Communication-Computation Trade-off}
We investigate the trade-off between on-device computation workload and communication overhead for different edge inference schemes
in Fig. \ref{fig:communication-computation trade-off}. Different points at a trade-off curve show the results of different model split points.
The split points with more than 1\% loss of inference accuracy compared to the original ResNet-50 are excluded.
It is observed that the proposed AutoML framework achieves up to 87.5\% communication overhead reduction compared to server-based 
inference and saves up to 70.4\% of on-device computations compared to simple model partition. Besides, there are more points on 
the trade-off curve of the proposed framework with both smaller on-device computation workload and communication overhead than those of the 
baselines, which demonstrates the advantages of applying DRL algorithms in searching for the optimal on-device model sparsification level and 
intermediate feature compression ratio for device-edge co-inference.
In addition, we see from Fig. \ref{fig:communication-computation trade-off} that all device-edge co-inference schemes show negligible 
effect in reducing the communication overhead when the split point is selected at early network layers. This is because few network layers is 
insufficient to extract low-entropy feature vectors of the raw data. 

\subsection{End-to-end Inference Latency}
We choose the Raspberry Pi 3B+ and the Honor 8 Lite smartphone as mobile devices, and deploy an edge server with 
a GTX 1080Ti graphics processing unit (GPU) to implement the edge inference schemes using Tensorflow Lite. The end-to-end inference latency 
includes both the computation and communication latency. We measure the computation latency at the mobile devices and the edge server, 
and calculate the communication latency as the ratio between the size of the intermediate feature vector and the communication 
rate. As an example, we select the split point right after the \textit{Conv\_4x} unit \cite{b18} in ResNet-50.

Fig. \ref{fig:latency analysis} shows the inference latency of different edge inference schemes by varying the communication rate between the
mobile device and the edge server. From both figures, it is observed that the proposed 
framework achieves lower end-to-end inference latency compared to other device-edge co-inference schemes, which again validates its 
effectiveness. Nevertheless, when the communication rate is below (above) a certain thresold, on-device inference (server-based inference) 
results in smaller latency, showing that the selection of model split point should be adaptive to the wireless environments. Besides, for a 
given inference latency requirement, mobile devices with stronger computation capability (i.e., Honor 8 Lite) poses less stringent 
requirement on the communication bandwidth. This demonstrates the importance of a wise choice of mobile devices in edge intelligent systems, 
which should balance cost and performance. 

\section{Conclusions} \label{sec:conclusions}
In this paper, we proposed an AutoML framework for communication-computation efficient device-edge co-inference. The proposed framework 
utilizes DRL to determine the optimal model sparsity level and intermediate feature compression ratio in order to reduce both the on-device 
computation workload and communication overhead. Experiment results show the competence of the proposed framework in achieving a better 
communication-computation trade-off and lower end-to-end inference latency. In the future, we will extend the proposed AutoML framework for
energy-efficient device-edge co-inference.

\end{document}